\documentclass[lettersize,conference]{IEEEtran}
\usepackage{amsmath,amsfonts}
\usepackage{algorithmic}
\usepackage{array}
\usepackage[caption=false,font=normalsize,labelfont=sf,textfont=sf]{subfig}
\usepackage{textcomp}
\usepackage{stfloats}
\usepackage{url}
\usepackage{verbatim}
\usepackage{graphicx}

\usepackage{nicematrix}
\usepackage{algorithm}
\usepackage{amssymb}
\usepackage{xcolor}
\usepackage{booktabs}
\usepackage{multirow}
\usepackage{adjustbox}
\usepackage{hyperref}

\usepackage{cellspace}
\newcommand{\fst}[1]{\color{red}{\textbf{#1}}}
\newcommand{\snd}[1]{\color{blue}{#1}}
\newcommand{\TwoRow}[1]{\multirow{2}{*}{#1}}
\newcommand{\NewDiffusionModel}{State Propagation Diffusion Model}
\def\NewDM{SPDM}

\def\BibTeX{{\rm B\kern-.05em{\sc i\kern-.025em b}\kern-.08em
    T\kern-.1667em\lower.7ex\hbox{E}\kern-.125emX}}
\usepackage{balance}
\begin{document}

\title{TrajWeaver: Trajectory Recovery with \NewDiffusionModel}



\makeatletter
\newcommand{\linebreakand}{%
  \end{@IEEEauthorhalign}
  \hfill\mbox{}\par
  \mbox{}\hfill\begin{@IEEEauthorhalign}
}
\makeatother

\author{\IEEEauthorblockN{Jinming Wang}
\IEEEauthorblockA{University of Exeter\\jw1294@exeter.ac.uk}
\and
\IEEEauthorblockN{Hai Wang}
\IEEEauthorblockA{Southeast University \& JD Logistic\\hai@seu.edu.cn}
\and
\IEEEauthorblockN{Hongkai Wen}
\IEEEauthorblockA{University of Warwick\\hongkai.wen@warwick.ac.uk}
\linebreakand
\IEEEauthorblockN{Geyong Min}
\IEEEauthorblockA{University of Exeter\\G.Min@exeter.ac.uk}
\and
\IEEEauthorblockN{Man Luo}
\IEEEauthorblockA{University of Exeter\\M.Luo@exeter.ac.uk}}


\maketitle

\begin{abstract}
With the proliferation of location-aware devices, large amount of trajectories have been generated when agents such as people, vehicles and goods flow around the urban environment. These raw trajectories, typically collected from various sources such as GPS in cars, personal mobile devices, and public transport, are often \textit{sparse} and \textit{fragmented} due to limited sampling rates, infrastructure coverage and data loss. In this context, trajectory recovery aims to reconstruct such sparse raw trajectories into their dense and continuous counterparts, so that fine-grained movement of agents across space and time can be captured faithfully. Existing trajectory recovery approaches typically rely on the prior knowledge of travel mode or motion patterns, and often fail in densely populated urban areas where accurate maps are absent. In this paper, we present a new recovery framework called TrajWeaver based on probabilistic diffusion models, which is able to recover dense and refined trajectories from the sparse raw ones, conditioned on various auxiliary features such as Areas of Interest along the way, user identity and waybill information. The core of TrajWeaver is a novel State Propagation Diffusion Model (SPDM), which introduces a new state propagation mechanism on top of the standard diffusion models, so that knowledge computed in earlier diffusion steps can be reused later, improving the recovery performance while reducing the number of steps needed. Extensive experiments show that the proposed TrajWeaver can recover from raw trajectories of various lengths, sparsity levels and heterogeneous travel modes, and outperform the state-of-the-art baselines significantly in recovery accuracy. Our code is available at: \url{https://anonymous.4open.science/r/TrajWeaver/}
\end{abstract}


\section{Introduction}
\IEEEPARstart{T}{he} ubiquity of GPS-enabled devices has revolutionized urban data collection, producing vast amounts of trajectory data that serve as the foundation for data-driven smart city applications. For instance, trajectories can be used to discover road networks \cite{Smallmap}, classify vehicle types \cite{CNN_vehicle_cls} and identify transportation mode \cite{trans_mode_identify}. However, the collected trajectories are often sparse and disjointed due to constraints such as limited sampling rates and intermittent signal loss. This sparsity poses a significant challenge in accurately capturing the continuous movement of agents across urban landscapes. Trajectory recovery emerges as an essential task, aiming to reconstruct these incomplete trajectories into detailed and seamless paths that accurately reflect the agents' movements. Specifically, given a sparse trajectory and various additional contexts such as user ID or waybill data, trajectory recovery is to generate and fill points into the gaps within the sparse trajectory. However, this task is fraught with challenges. First, the moving agents can exhibit heterogeneous mobility patterns. Especially in last-mile delivery scenarios, the couriers can frequently enter or exit apartments, moving quickly or staying at one location for a long time. The second challenge involves the diverse lengths and sparsity levels of trajectories, which requires the recovery model to have good scalability. Lastly, the detailed map information is usually absent in complex metropolitan or residential areas. Most existing methods perform poorly in these scenarios, struggling to generalize across diverse transportation modes or densely populated areas.

\begin{figure}[t]
\centerline{\includegraphics[width=0.99\linewidth]{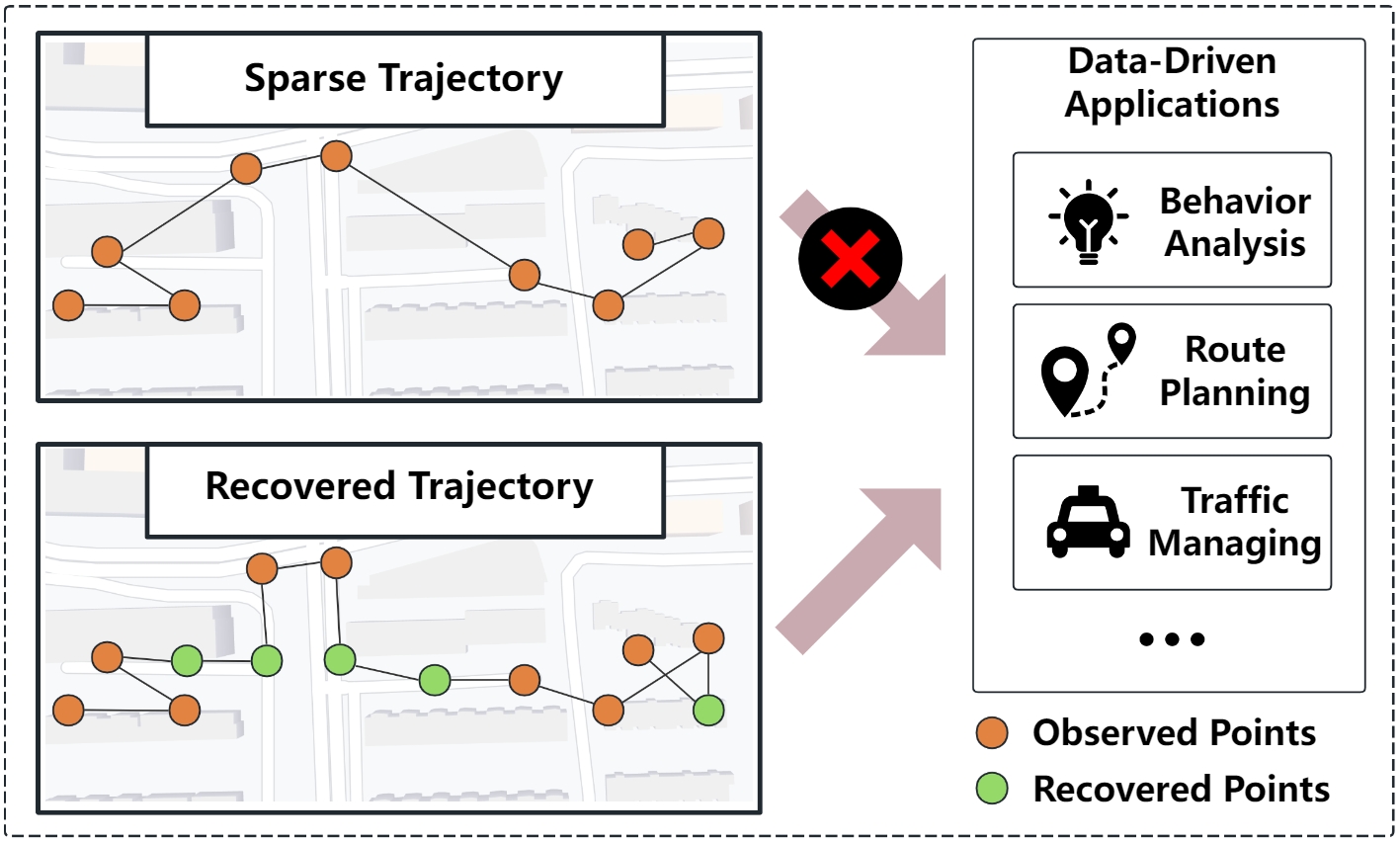}}
\caption{The collected trajectory is too sparse to be used in data-driven applications.}
\label{fig:UnformatTraj}
\end{figure}

Recently, diffusion models have demonstrated robust performance in content recovery tasks such as image inpainting ~\cite{ImageInpainting} and time series imputation ~\cite{PriSTI}. A diffusion model operates through two multi-step processes. The diffusion process corrupts the data through recursive noise injection, eventually resulting in pure noise. Conversely, the denoising process employs neural networks to progressively remove the noise at each step, thereby recovering the original data. Compared to other content generation methods like Variational Autoencoders (VAEs) ~\cite{VAE, TrajVAE} and Generative Adversarial Networks (GANs) ~\cite{GAN, TrajGANContinuous}, diffusion models exhibit superior modeling, pattern capturing and content generation capabilities across a range of tasks ~\cite{ddpm_beat_gan_topology, text-to-image-vqvae, Sora}. These strengths suggest that diffusion models hold significant potential for enhancing trajectory recovery, offering more accurate and robust reconstructions of trajectories.

Although the unique multi-step procedure endows diffusion models with strong recovery ability, several challenges arise when applying them to trajectory recovery. The first challenge lies in integrating diverse recovery conditions into diffusion models in an economic and efficient way. Different from trajectory generation task which generates the entire trajectory based on few conditions, trajectory recovery generates additional points given various conditions. In addition to the sparse trajectory being the foundation of this recovery task, extra contexts include the prior knowledge such as linear interpolated points, and numerous additional contexts such as user ID, weekday, and waybill information. Given that diffusion models perform recovery through many denoising steps, where conditions are fused at each step, the integration of conditions must be multi-modal, computationally efficient, and sensitive to the spatio-temporal correspondence among these conditions. A common way of conditioning in diffusion models is through cross-attention, as stated in \cite{StableDiffusion}, which employs a pre-trained module to integrate and convert conditions into an embedding, then fuse the embedding into the denoising neural network. However, due to the squared memory usage of cross-attention modules, this way of condition fusing is not efficient, especially for long trajectories.

As the basic version of diffusion model, Denoising Diffusion Probabilistic Model (DDPM) ~\cite{DDPM} often takes thousands of denoising steps to complete recovery. Techniques such as Denoising Diffusion Implicit Model (DDIM) ~\cite{DDIM}, DPM Solver ~\cite{DPMSolver} and Pseudo Numerical Methods for Diffusion Models (PNDM) ~\cite{PNDM} can reduce the number of steps needed to tens, while other methods have accelerated the denoising process through model pruning and distillation ~\cite{DiffPruning, one_step_diffusion, DeepCache}. However, most approaches sacrifice recovery quality to achieve lower latency, and it is challenging to simultaneously ensure fast and high-quality recovery. The core issue that prevents efficient recovery is the inherently isolated nature of the denoising steps. The denoising process often takes hundreds of steps, where each step performs extensive feature extraction. However, as defined by diffusion framework, the denoised content is the only information passed to the next step, which hinders the sharing of knowledge among steps. A more efficient recovery process could be realized by promoting greater collaboration among denoising steps, enabling the reuse of features, and thereby achieving high-quality recovery with significantly fewer steps.

Scalability is a critical challenge in applying diffusion models to trajectory recovery, given the varying lengths and sparsity levels of trajectories. In logistics scenarios, trajectory lengths can range from tens of points to thousands, and the total distance can vary from hundred meters to kilometers. Short trajectories often lack sufficient conditional information, which requires the model to accurately reconstruct them with limited information. In the case of sparse trajectories, more points should be generated and inserted into the trajectory, so diffusion models introduce more noise into the data and leads to greater instability. On the other hand, long trajectories require the model to distill useful information at each denoising step, which can increase model complexity and slow down recovery. Moreover, the diverse spatio-temporal patterns correspond to different trajectory length and sparsity can further complicate the condition fusion process. Therefore, ensuring the scalability of diffusion models is essential for their successful application in trajectory recovery.

To address the challenges outlined above, we propose a novel diffusion-based trajectory recovery framework called TrajWeaver, which is capable of effectively aggregate and fuse diverse conditions into the multi-step recovery process. To achieve high-quality and low-latency trajectory recovery simultaneously, TrajWeaver introduces a new variant of the diffusion model, \NewDiffusionModel{} (\NewDM{}). In particular, it implements a state propagation pipeline on top of the standard denoising process, which allows knowledge sharing and features reuse among steps. This inter-step information sharing not only improves recovery efficiency, but also enhances scalability by transmitting multi-scale features within state. The design of TrajWeaver is anchored in three key elements: an advanced module for effective condition aggregation, a neural network that enables state fusing and propagation, and a specialized training algorithm tailored for \NewDM{}. Through extensive experiments, TrajWeaver demonstrates its ability to recover both pedestrian and vehicle trajectories in complex urban environments with high accuracy and scalability.

Overall, our contributions can be summarized as follows:

\begin{itemize}
\item We propose a novel diffusion-based trajectory recovery framework named TrajWeaver, which is capable of aggregating various forms of conditions to achieve fast and accurate trajectory recovery in complex environments with dynamic moving patterns.
\item We introduce \NewDM{}, which consists a new state propagation mechanism on top of the standard diffusion models, so that more efficient recovery process and better scalability can be achieved.
\item We conduct extensive experiments to validate the effectiveness of the proposed method, including comparisons with existing methods, ablation studies and use case analysis.
\end{itemize}

\section{Related Work}\label{sec:related_works}
\noindent\textbf{Trajectory Recovery.}
Numerous efforts have been dedicated to increase the accuracy of trajectory recovery. Certain map-based methods ~\cite{HumanTrajCompletionTrans, TaxiTrajRec, MTrajRec, RNTrajRec} utilize a predefined set of points of interest (POI) or a pre-collected road network as strong prior knowledge, but the detailed map information is not always guaranteed. In contrast, DHTR ~\cite{TrajRecCalibKF} performs free space trajectory recovery leveraging RNNs, which does not require road network information. However, for very long and sparse trajectories, RNNs struggle to capture dependencies among points. This is because RNNs recovery consecutive points in auto-regressive manner, which may accumulate the recovery errors. AttnMove ~\cite{AttnMove} is a method based on Transformers ~\cite{Transformer}, which excel in capturing long-term dependencies within sequence. In particular, it assumes that the agent moves in periodic manner, thus similar trajectories will be produced in each period. By aggregating all history trajectories and the current trajectory via inter- and intra-trajectory attention, the underlying mobility patterns of the agent can be revealed. TrajBERT ~\cite{TrajBert} is another Transformer-based method that is mainly built on BERT \cite{BERT}, which incorporates multi-aspect spatial-temporal aware designs.

\noindent\textbf{Diffusion-based Trajectory Generation.}
There are no existing works that apply diffusion models to trajectory recovery, but diffusion-based methods on time series recovery and trajectory generation can be used as references. Existing approaches have successfully applied diffusion models to generate and recover time series data, including CSDI ~\cite{CSDI} and SSSD ~\cite{SSSD}. In particular, PriSTI ~\cite{PriSTI} is a recent diffusion-based method designed for time series recovery, it utilizes spatio-temporal conditions and geographical factors for more accurate recovery. On the other hand, another thread of existing work uses diffusion models for trajectory generation rather than recovery. The main difference between these two tasks is that trajectory generation produces the full trajectory without knowing existing portions, and the aim of this task is usually to ensure the generated dataset having similar distribution as the original trajectory dataset. For example, DiffTraj ~\cite{DiffTraj} employs UNet ~\cite{UNet} to perform denoising steps with several simple contexts such as trajectory length, beginning point and end point. Diff-RNTraj ~\cite{diff_rn_traj} combines the advantages of both PriSTI and DiffTraj to perform trajectory generation with road network integrated.

\noindent\textbf{More Efficient Diffusion Models.}
Denoising Diffusion Probabilistic Models (DDPM) \cite{DDPM} is the basic version of diffusion model, which requires hundreds of denoising steps to complete recovery. Many techniques were proposed to reduce the redundancy and improve efficiency of DDPM. Such works include Denoising Diffusion Implicit Model (DDIM) \cite{DDIM} and DPM Solver \cite{DPMSolver}, which can shrink the denoising process to 10 to 20 steps with a cost of around 10\% to 30\% loss in recovery quality. There exists methods capable of distilling denoising process to one step \cite{one_step_diffusion}, but it sacrifices more result quality for extremely fast recovery. Pseudo Numerical Methods for Diffusion Models (PNDM) ~\cite{PNDM} is a more advanced diffusion sampler that performs fast denoising process while preserves the original result quality. Moreover, model compression and quantization methods such as Diff-Pruning \cite{DiffPruning} and PTQD \cite{PTQD} can effectively reduce computation waste thus accelerate each denoising step. DeepCache \cite{DeepCache} is the most similar method to our approach, which optimizes denoising process by excluding certain parts of the UNet \cite{UNet} and reusing previously extracted features.

\begin{figure*}[t]
  \begin{center}
  \includegraphics[width=0.99\textwidth]{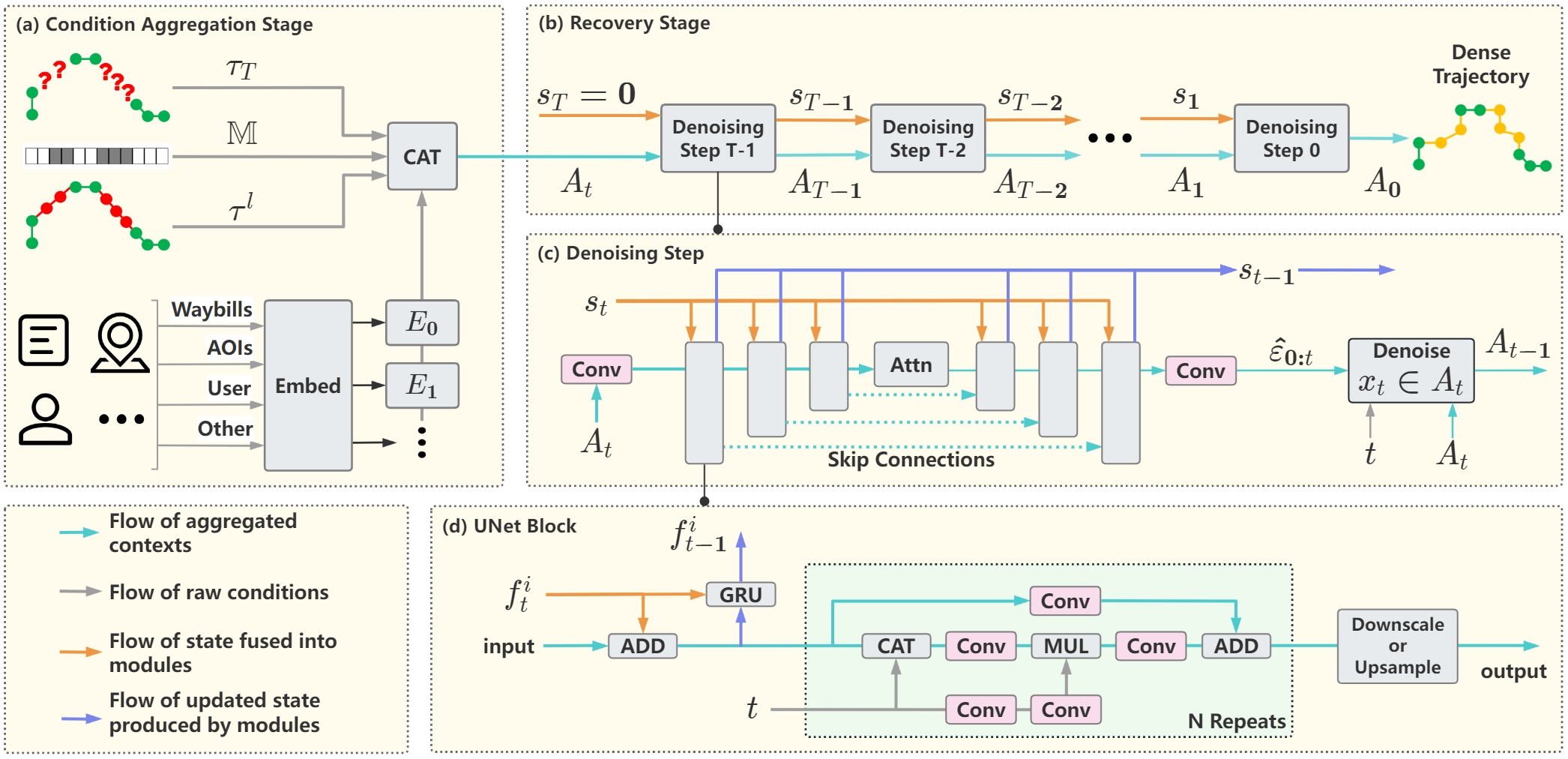}
  \caption{TrajWeaver Framework. (a): The aggregation of trajectory, prior knowledge and all other contexts. (b): The denoising pipeline for trajectory recovery. (c): The denoising step, including the proposed neural network architecture. (d): A single block in the proposed network.}
  \label{fig:Main}
  \end{center}
\end{figure*}

\section{Preliminaries}\label{sec:backgrounds}
\subsection{Problem Formulation}

\noindent\textbf{Trajectory:} A trajectory $\tau=\{p_0, p_1, ...\}$ is a chronologically ordered sequence that describes the movement of an entity, where each element is a spatio-temporal point $p_i=(lng_i, lat_i, time_i)$ consisting of longitude, latitude and time stamp.

\noindent\textbf{Sparse Trajectory:} Given a dense trajectory $\tau$, a sparse trajectory $\tilde{\tau}$ contains a subset of $\tau$ and fails to accurately describe the continuous movement of the entity in the real world.

\noindent\textbf{Query:} A query $Q$ is a manually selected or procedurally generated sequence of time stamps. We want to determine the location of the moving entity at each specified time stamp in $Q$.

\noindent\textbf{Trajectory Recovery:} Given sparse trajectory $\tilde{\tau}$, query $Q$, and other types of contexts such as date, weekday, user ID, waybills, etc. The goal of trajectory recovery is to accurately predict the unknown locations correspond to the provided $Q$. 

\subsection{Diffusion Models}

\noindent\textbf{Diffusion Process.}
The diffusion process consists of a series of noise addition steps, where each step is formulated as in Equation \ref{eq:x_t_to_tp1}. In this equation, $t \in [0, T)$ denotes the index for the diffusion step, $\alpha_t$ and $\beta_t$ are noise schedules and $\alpha_t = 1 - \beta_t$. $\varepsilon_{t:t+1}$ represents single step noise drawn from $\mathcal{N}(0, 1)$.
\begin{equation}\label{eq:x_t_to_tp1}
x_{t+1}=\sqrt{\alpha_t}x_{t}+\sqrt{\beta_t}\varepsilon_{t:t+1}
\end{equation}

We can further derive and compute $x_t$ directly from the original content $x_0$ instead of applying the forward step $t$ times, as shown in Equation \ref{eq:x_0_to_tp1}. Here, $\bar{\alpha}_t=\prod_{i=1}^t \alpha_i$, and $\varepsilon_{0:t+1}\sim\mathcal{N}(0,1)$ represents the multi-step noise added to $x_0$ to produce $x_t$ in one step. The notation 

\begin{equation}\label{eq:x_0_to_tp1}
x_{t+1}=\sqrt{\Bar{\alpha}_t}x_0+\sqrt{1-\Bar{\alpha}_t}\varepsilon_{0:t+1} 
\end{equation}

\noindent\textbf{Denoising Process.}
The denoising process aims to reconstruct clear content $x_0$ from random noise $x_T$ through a series of denoising steps. Given noisy content $x_{t+1}$ at step $t+1$ where $t\in[0,T-1]$, the aim of this denoising step is to predict the mean and standard deviation of the less noisy content $x_{t}$. The equation below explains one denoising step.
  
  \begin{equation}\label{eq:back_mu_std}
  \begin{aligned}
      \mu_{t}&=\frac{x_{t+1}-\beta_t * \varepsilon^{pred}_{0:t+1}}{\sqrt{\Bar{\alpha}_t}} \\
      \sigma_{t}&=\sqrt{\frac{1-\Bar{\alpha}_{t-1}}{1-\Bar{\alpha}_t}*\beta_t} * z, \quad z \sim \mathcal{N}(0, 1) \\
      x_{t}&=\mu_{t}+\sigma_{t}
  \end{aligned}
  \end{equation}

A neural network is trained to predict $\varepsilon_{0:t+1}$ in equation \ref{eq:back_mu_std} since it is the only unknown value. During training, we choose $t$ from valid range from 0 to $T-1$, then sample $\varepsilon_{0:t} \sim \mathcal{N}(0, 1)$ and apply diffusion forward process to $x_0$ to obtain $x_t$, where $x_0$ is the original content and $x_t$ is the noisy content. The model is trained to produce $\epsilon^{pred}_{0:t}$ and the loss is computed using mean squared error.

\section{Methodology}\label{sec:methods}
\subsection{TrajWeaver Overview}\label{sec:Method_Overview}

  \noindent\textbf{Condition Aggregation Stage.} Before entering the pipeline of TrajWeaver, we first need to define all input components that are crucial for the model's performance. The main input is a sparse trajectory $\tilde{\tau}$, which represents a sequence of spatio-temporal points sampled at irregular intervals. Alongside this sparse trajectory, a query $Q$ is selected either manually or procedurally, containing a series of time stamps at which the locations need to be predicted. These time stamps from $Q$ are inserted into the sparse trajectory $\tilde{\tau}$, with the corresponding unknown locations initialized with random noise values, denoted as $x_T$. The outcome of this initialization is the trajectory $\tau_T$, where TrajWeaver's objective is to convert the noisy locations $x_T \in \tau_T$ into accurate predictions $x_0 \in \tau_0$, representing the original, noise-free trajectory. The process of composing $\tau_T$ is depicted in Figure \ref{fig:decomp_traj}.

  To facilitate the recovery process, we generate a binary mask $\mathbb{M}$, where each entry is set to 0 or 1 to indicate whether a point is part of the original trajectory or an inserted point, respectively. This mask helps the model distinguish between observed and unobserved data points. Furthermore, to incorporate prior spatio-temporal knowledge into the model, we create an additional trajectory representation, $\tau^l$, where unknown locations are filled with linearly interpolated values instead of random noise. This serves as a preliminary estimate or "prior guess" of the true trajectory, providing a useful baseline for the model to build upon during the diffusion process.

  \begin{figure}
      \centering
      \includegraphics[width=0.8\linewidth]{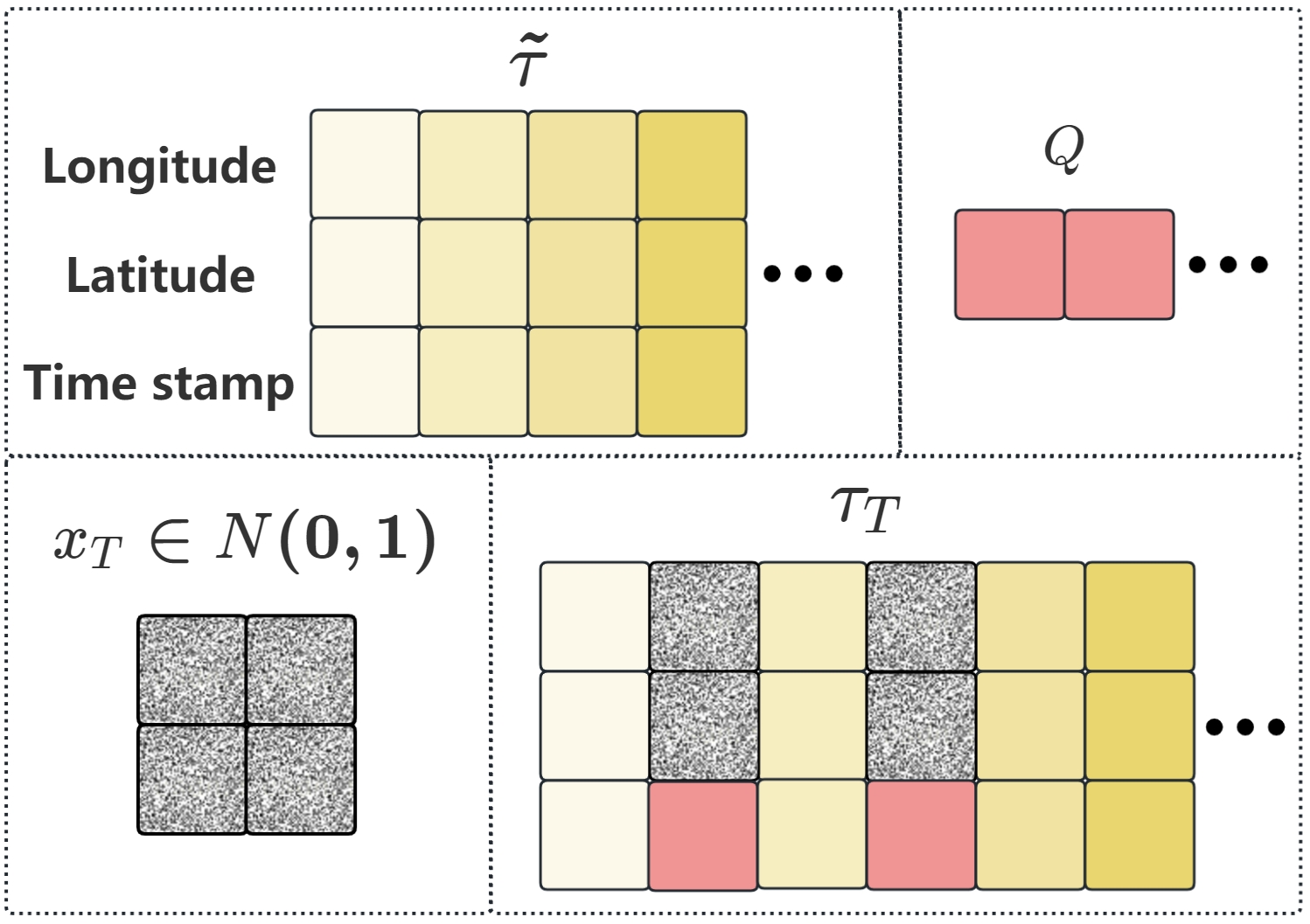}
      \caption{The trajectory with noise added $\tau_T$ is a composition of the given sparse trajectory $\tilde{\tau}$, the query $Q$ and the noise $x_T$.}
      \label{fig:decomp_traj}
  \end{figure}

  In addition to the previous inputs, various contextual features ${C_0, C_1, ...}$ that are associated with the trajectory are also considered. Due to the sequential form of trajectories, the positions of tokens contain important spatio-temporal knowledge. To ensure good positional correspondence among conditions, we also want to process the contexts ${C_0, C_1, ...}$ to sequential forms just like other conditions. However, the forms of these contexts can vary widely in different industrial scenarios, a universal embedding module design is inappropriate. Therefore, we assume a specifically designed module $Embed_i$ for each context $C_i$, which processes it into a sequential embedding $E_i$ of the same length as $\tau_T$. These modules can be as trivial as several convolutional layers or fully-connected layers. All conditions are now in sequential format with the same length. Instead of applying cross attention to fuse them as in Stable Diffusion \cite{StableDiffusion}, we simply concatenate them together, witch keeps positional alignment among sequences, thus ensures good spatio-temporal correspondence. The final aggregated condition is denoted as $A_T$. This stage is illustrated in Figure \ref{fig:Main} (a) and formulated as follows:

  \begin{equation} \label{eq:embed_stage}
  \begin{aligned}
      A_T &\gets Concat(\tau_T, \tau^l, \mathbb{M}, E_0, E_1, ...)\\
      E_i &\gets Embed_i(C_i)
  \end{aligned}
  \end{equation}

  \noindent\textbf{Recovery Stage.} TrajWeaver recovery stage focuses on removing noise from $x_T \in A_T$, while all other portions of $A_T$ remain fixed throughout the recovery procedure. As shown in Figure \ref{fig:Main} (b), recovery takes $T$ steps counting down from $T-1$ to 0. Each denoising step is illustrated in Figure \ref{fig:Main} (c). Given $A_t$ in step $t$, a neural network is utilized to make a prediction $\hat{\varepsilon}_{0:t}$. Then, $x_t$ is denoised to $x_{t-1}$, which also transforms $A_t$ to $A_{t-1}$. A typical denoising step in diffusion models takes only $A_t$ and $t$ as inputs then produces $\hat{\varepsilon}_{0:t}$. As a result, the traditional denoising steps are nearly isolated, and some similar features can be repeated extracted in each step. This significant redundancy makes the denoising process inefficient, and prevents the model from achieving low latency and high recovery quality simultaneously. Therefore, TrajWeaver employs \NewDM{} which allows inter-step information sharing and feature reuse. Specifically, as indicated by the orange arrows in Figure \ref{fig:Main} (b), each denoising step produces an additional state $s_{t-1}$ that contains useful information for subsequent steps, and each step also absorbs state $s_t$ from the previous step, with the initial state $s_T$ filled with 0. This state propagation mechanism is completed by the specially designed denoising neural network, as formulated in Equation \ref{eq:LDDM_step}.

  \begin{equation}\label{eq:LDDM_step}
    \hat{\varepsilon}_{0:t}, s_{t-1} = \mathbf{Net}(A_t, t, s_t)
  \end{equation}

\subsection{Network Architecture} \label{sec:LDDM}

  As illustrated in Figure \ref{fig:Main} (c), the denoising neural network is based on UNet \cite{UNet} architecture which facilitates multi-scale feature extraction and enhances scalability. Each UNet block is shown in Figure \ref{fig:Main} (d) Multi-head self-attention modules are employed in the middle of the network, allowing global receptive field and better capture spatio-temporal dependencies in long and sparse trajectories. The unique design within TrajWeaver denoising neural network is the implementation of state propagation mechanism, which requires solving three key problems: determining the format of states, fusing state $s_{t}$ into the neural network, and updating $s_t$ to produce $s_{t-1}$.

  \noindent\textbf{Format of State.} We design the state as multiple feature sequences, denoted as $s_t = \{f_t^0, f_t^1, f_t^2, ...\}$. This state representation has two advantages. First, since trajectory features are all in sequential format, using the same format for state features can ensure a strong positional correspondence during feature fusing. While other feature formats like vectors do not have this property, it also requires additional processing to fuse vectors with sequences. Second, the dynamic lengths and sparsity levels of trajectories can affect the patterns within the input contexts at different scale. We make $s_t$ multiple sequences with different scales to align with the UNet's multi-level design, allowing better scalability.

  \noindent\textbf{Fusing Previous State.} Each feature sequence within $s_t$ is fused into each UNet block as shown in Figure \ref{fig:Main} (d). There are several ways for sequence fusing, cross-attention is a widely adopted choice for multi-modal context fusion in diffusion models. However, the multi-scale sequential form of features have already ensure the good positional correspondence, which makes simple addition or concatenation sufficient for state fusion. In the figure, we show state feature fusion using addition.

  \noindent\textbf{Propagating State.} To update $s_t$ to $s_{t-1}$, a GRU cell is employed to propagate each $f_t^i$ to $f_{t-1}^i$. By adopting the GRU cell, $s_{t-1}$ not only encapsulates knowledge in the current denoising step, but also contains knowledge of all previous steps, which allows long-range inter-step information sharing. Ideally, at the last few steps of the denoising process, the state will contain the most useful knowledge extracted in the previous tens to hundreds of steps.

  \subsection{Training of TrajWeaver}\label{sec:LDDM_train_practice}

  During TrajWeaver training, a dense ground truth trajectory $\tau_0$ is provided, some time stamps are randomly selected as the query $Q$, and the corresponding locations are $x_0$. The diffusion process adds $T$ steps of noise to $x_0$, producing $x_T$. Since $x_0$ is part of $\tau_0$, applying diffusion to $x_0$ also transforms $\tau_0$ into $\tau_T$.

  The training of \NewDM{} differs significantly from typical diffusion model training, because the state propagation introduces more dependency among denoising steps. As suggested in Equation \ref{eq:LDDM_step} and Figure \ref{fig:Main} (b), denoising step $t$ cannot be performed without $s_{t+1}$ from previous step $t+1$. Consequently, unlike in typical diffusion model training where a random $t$ is chosen in each training sample, we must iterate over $t$ from $T-1$ to $0$ to ensure that we always have previous state $s_{t+1}$, with the initial state $s_T$ set to zero tensors.

  Another problem is to ensure $s_t$ containing useful features for later steps. Since the $s_t$ is produced by the GRU cell and is used in the next denoising step, the parameters of GRU cannot be optimized when each step is trained independently. Therefore, we have to train multiple steps jointly in each training iteration. The ideal approach is to train the denoising process as a whole, thus the overall objective of \NewDM{} is presented in Equation \ref{eq:arch_obj}, where $\theta$ represents the learnable parameters of the model.

  \begin{equation}\label{eq:arch_obj}
      \min_{\theta}\sum_{t=1}^T MSE(\hat{\varepsilon}_{0:t}, \varepsilon_{0:t})
  \end{equation}

  However, it is unrealistic to train all tens to hundreds of steps in each iteration. Therefore, we split the denoising process into smaller segments consisting of several adjacent steps. The new objective for training 2 steps in a single iteration is formulated in Equation \ref{eq:arch_obj_2step}.
  
  \begin{equation}\label{eq:arch_obj_2step}
      \min_{\theta} (MSE(\hat{\varepsilon}_{0:t+1}, \varepsilon_{0:t+1}) + MSE(\hat{\varepsilon}_{0:t}, \varepsilon_{0:t}))
  \end{equation}

  Standard diffusion model samples $\varepsilon_{0:t} \sim \mathcal{N}(0, 1)$ independently in each training iteration. This becomes inappropriate when multiple steps are included in one iteration, because $\varepsilon_{0:t+1}$ and $\varepsilon_{0:t}$ are dependent. In fact, $\varepsilon_{0:t+1}$ depends on all previous single-step noises $\{\varepsilon_{0:1}, \varepsilon_{1:2}, ..., \varepsilon_{t:t+1}\}$ and multi-step noises $\{\varepsilon_{0:1}, \varepsilon_{0:2}, ..., \varepsilon_{0:t}\}$. Therefore, $\varepsilon_{0:t}$ and $\varepsilon_{0:t+1}$ cannot be sampled directly from Gaussian distribution. Nonetheless, the single-step noises $\varepsilon_{t:t+1}$ remain independent, which allows us to sample all single-step noises $\{\varepsilon_{t:t+1} | t \in [0, T)\}$ and then derive multi-step noises using Equation \ref{eq:noise_propagate}:

  First, we combine equation \ref{eq:x_t_to_tp1} and \ref{eq:x_0_to_tp1}
  \begin{equation*}\label{eq:1_eq_2}
      \sqrt{\Bar{\alpha}_t}x_0+\sqrt{1-\Bar{\alpha}_t}\varepsilon_{0:t+1} = x_{t+1}=\sqrt{\alpha_t}x_{t}+\sqrt{\beta_t}\varepsilon_{t:t+1}
  \end{equation*}

  We can also get $x_t$ from $x_0$ with equation \ref{eq:x_0_to_tp1}, and then replace $x_t$ with a function of $x_0$ and $\varepsilon_{0:t}$.
  \begin{equation*}\label{eq:noise_propagate}
  \begin{aligned}
      \sqrt{\Bar{\alpha}_t}x_0+\sqrt{1-\Bar{\alpha}_t}\varepsilon_{0:t+1}&=\sqrt{\alpha_t}(\sqrt{\Bar{\alpha}_{t-1}}x_0+\sqrt{1-\Bar{\alpha}_{t-1}}\varepsilon_{0:t}) \\
      &\quad +\sqrt{\beta_t}\varepsilon_{t:t+1} \\
      \sqrt{1-\Bar{\alpha}_t}\varepsilon_{0:t+1}&=\sqrt{\alpha_t}\sqrt{1-\Bar{\alpha}_{t-1}}\varepsilon_{0:t}+\sqrt{\beta_t}\varepsilon_{t:t+1} \\
      \varepsilon_{0:t+1}&=\frac{\sqrt{\alpha_t}\sqrt{1-\Bar{\alpha}_{t-1}}\varepsilon_{0:t}+\sqrt{\beta_t}\varepsilon_{t:t+1}}{\sqrt{1-\Bar{\alpha}_t}}
  \end{aligned}
  \end{equation*}

  The resulting Equation \ref{eq:noise_propagate} shows a way to inference $\varepsilon_{0:t+1}$ based on $\varepsilon_{0:t}$ and $\varepsilon_{t:t+1}$. We call it a noise compose equation and simplify it as:

  \begin{equation}
      \varepsilon_{0:t} \gets Compose(\varepsilon_{0:t+1}, \varepsilon_{t:t+1})
  \end{equation}
  
  With this equation and randomly sampled list of single-step noises $\{\varepsilon_{0:1}, \varepsilon_{1:2}, ... \varepsilon_{T-1:T}\}$, we can obtain the list of multi-step noises through a recursive process shown below.
  
  \begin{equation*}
      \begin{aligned}
          \varepsilon_{0:2} &\gets Compose(\varepsilon_{0:1}, \varepsilon_{1:2})\\
          \varepsilon_{0:3} &\gets Compose(\varepsilon_{0:2}, \varepsilon_{2:3})\\
          \varepsilon_{0:4} &\gets Compose(\varepsilon_{0:3}, \varepsilon_{3:4})\\
          &...\\
          \varepsilon_{0:T} &\gets Compose(\varepsilon_{0:T-1}, \varepsilon_{T-1:T})\\
      \end{aligned}
  \end{equation*}

\begin{algorithm}[H]
\caption{Training One Sample (2-step)}\label{algo:training}
\begin{algorithmic}[1]
    \STATE Generate $\{\varepsilon_{t:t+1} | t \in [0, T)\}$
    \STATE Derive $\{\varepsilon_{0:t+1} | t \in [0, T)\}$
    \STATE Apply diffusion to $\tau_0$ to get $\{\tau_t | t \in [1, T]\}$
    \STATE $s_T \gets 0$
    \FOR{$t \gets T-1$ to $0$}

        \STATE $E_i \gets Embed(C_i)$ for additional contexts.

        \STATE $A_t, A_{t+1} \gets$ Concat $\tau_{t}$ and $\tau_{t+1}$ with $\tau^l$, $\mathbb{M}$, $E_i$ 

        \STATE $\hat{\varepsilon}_{0:t+1}, s_t \gets Net(A_{t+1}, s_{t+1}, t+1)$
        \STATE $\hat{\varepsilon}_{0:t}, s_{t-1} \gets Net(A_{t}, s_{t}, t)$
        \STATE $loss \gets MSE(\hat{\varepsilon}_{0:t+1}, \varepsilon_{0:t+1}) + MSE(\hat{\varepsilon_{0:t}}, \varepsilon_{0:t})$
        \STATE Perform Gradient Descent
    \ENDFOR
\end{algorithmic}
\end{algorithm}

The general principles of \NewDM{} training have been established, which makes TrajWeaver training possible. A naive procedure is presented in Algorithm \ref{algo:training}.

While the above design and formulas are sufficient for training, additional techniques are necessary in practical implementation. One major problem is that each trajectory is used $T$ training iterations and the counting down of $t$ applies to the entire batch of trajectories, illustrated as Figure \ref{fig:t_shared}. This shared countdown can lead to over-fitting to certain local range of $t$ and poor generalization across the entire range of $t$. As a result, the training is very unstable and the convergence extremely slow.

To mitigate this issue, a new batch managing design has to be adopted. We assign a different private $t$ for each sample in a mini-batch, as illustrated in Figure \ref{fig:t_consecutive}. To further enhance generalization over the entire denoising process and ensure stable convergence, we distribute $t$s uniformly within the range $[0, T-1]$ as shown in Figure \ref{fig:t_uniform}. The $t$ value for each sample is independently updated, and a new sample is loaded in-place when the $t$ for an old sample is reduced to 0.

\begin{figure}
\centering
\subfloat[Shared $t$ among all samples in a batch, all samples in the batch are reloaded together.\label{fig:t_shared}]{\includegraphics[width=\linewidth]{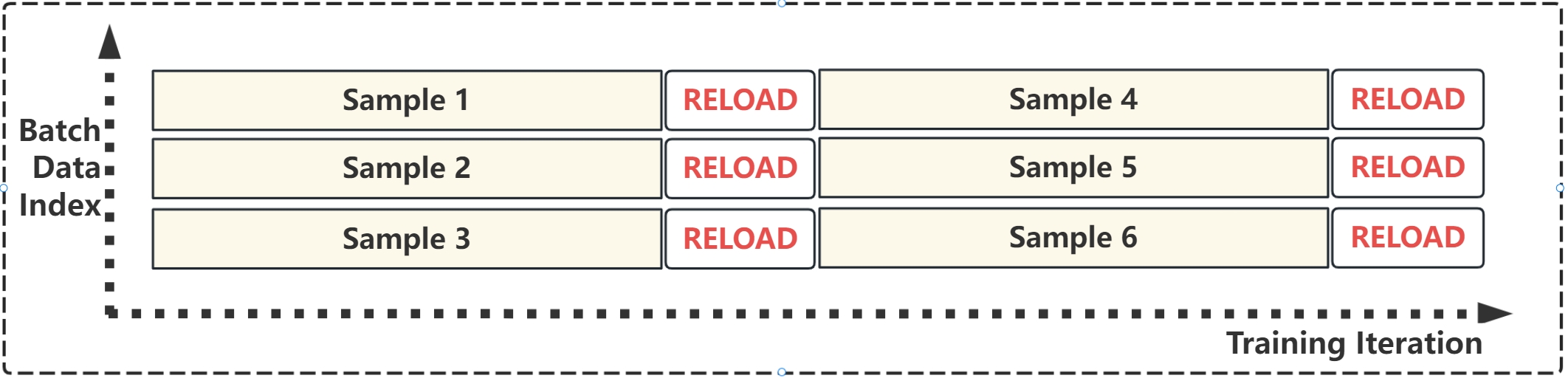}}\\
\subfloat[Consecutive $t$ for each sample, the samples in the batch are reloaded one after another.\label{fig:t_consecutive}]{\includegraphics[width=\linewidth]{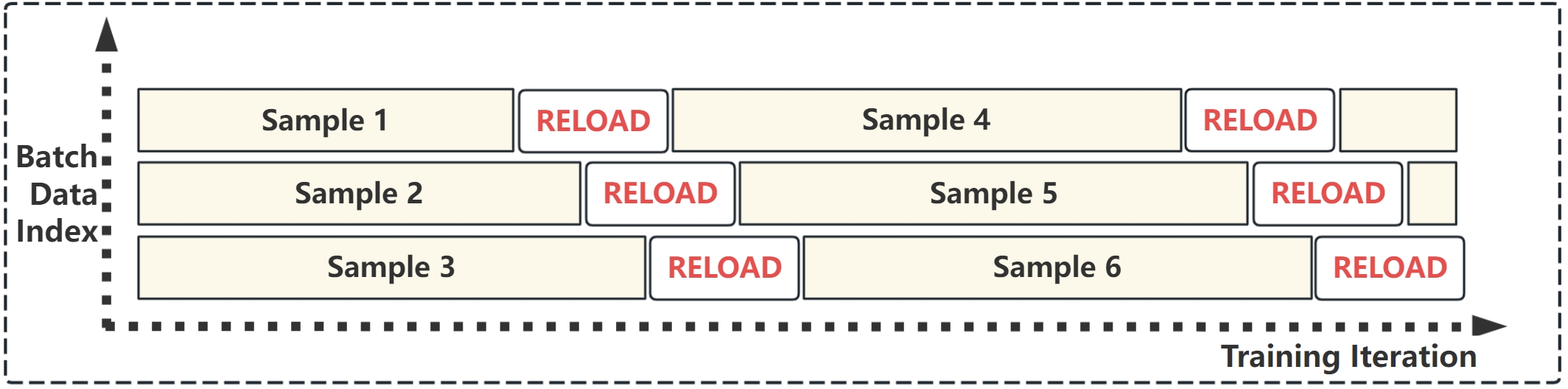}}\\
\subfloat[Uniformly distributed $t$ over the range $[0, T)$, the sample reloading is also uniformly distributed over time.\label{fig:t_uniform}]{\includegraphics[width=\linewidth]{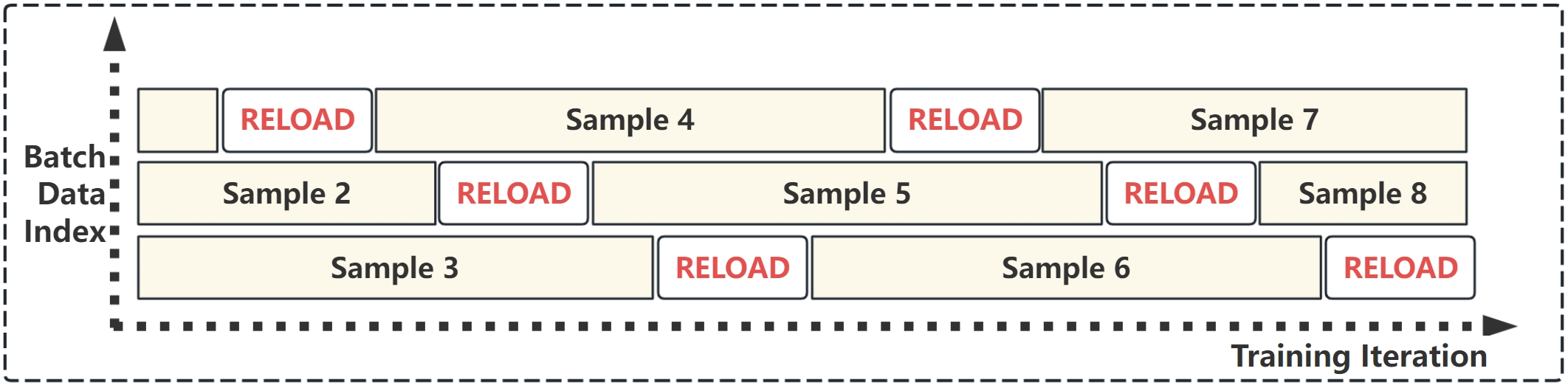}}
\caption{Three types of batch managing, the main difference is when to reload each batch sample.}
\label{fig:batch_loading}
\end{figure}

\section{Experiments}\label{sec:evaluation&comparison}
\begin{table}[t]
    \caption{Dataset Statistics After Pre-Processing}
    \label{tab:datasets}
    \centering
    \begin{tabular}{|l||c|c|c|}
        \hline
        \multirow{2}{*}{\textbf{Datasets}} & \multirow{2}{*}{\textbf{Logistics}} & \multirow{2}{*}{\textbf{Xi'an}} & \multirow{2}{*}{\textbf{Chengdu}} \\
        &&&\\
        \hline
        Duration & 4 months & 1 month & 1 month \\
        \hline
        \#Trajectories & 2584 & 131,123 & 114,110 \\
        \hline
        \#Points & 759,211 & 67,134,976 & 58,424,320 \\
        \hline
    \end{tabular}
  \end{table}

\subsection{Experimental Settings}

 \noindent\textbf{Datasets.} We use two datasets consisting of dense and smooth taxi trajectories collected in Xi'an city and ChengDu city, China. The third dataset is collected in real-world logistics scenarios generated by couriers on foot, this dataset involves couriers moving in apartment areas and exhibits irregular sample intervals. Table \ref{tab:datasets} shows the statistics of the three datasets.

  \begin{table}[t]
  \caption{The overall comparisons with sparsity of 0.5 and trajectory length of 512.}
  \label{tab:XianResult}
  \centering
  \begin{adjustbox}{max width=\textwidth}
    \setlength{\aboverulesep}{0pt}
    \setlength{\belowrulesep}{0pt}
  \begin{tabular}[t]{|l||l||c|c|c|}
    \hline
    \multirow{2}{*}{Metric} & \multirow{2}{*}{Method} & \multicolumn{3}{c|}{Dataset} \\
    \cmidrule{3-5}
    & & Xi'an & Chengdu & Logistics \\
    \hline
    MSE $\downarrow$
      & DeepMove & $5.297$ & $26.857$& $30.060$\\
      ($\times 10^{-3}$) & {AttnMove} & $0.068$ & $0.234$& $3.201$\\
      & {PriSTI} & $0.019$ & $0.292$& $2.206$\\
      & {DT + RP} & $0.326$ & $6.919$& $4.250$\\
      & {TW-DM} & \snd{$0.017$}& \snd{$0.202$}& \snd{$2.025$}\\
      & {TW} & \fst{$0.010$}& \fst{$0.159$}& \fst{$0.449$}\\
      \hline
    NDTW $\downarrow$
      & {DeepMove} & $36.295$ & $169.470$& $72.814$\\
      ($\times 10^{-3}$) & {AttnMove} & $2.257$ & $5.458$& $22.652$\\
      & {PriSTI} & $1.174$ & $3.975$& $12.656$\\
      & {DT + RP} & $8.473$ & $28.773$& $13.006$\\
      & {TW-DM} & \snd{$1.156$}& \snd{$3.809$}& \snd{$11.062$}\\
      & {TW} & \fst{$1.154$}& \fst{$3.597$}& \fst{$5.520$}\\
      \hline
    JSD $\downarrow$
      & {DeepMove} & $1.461$ & $1.786$& $2.913$\\
      ($\times 10^{-3}$) & {AttnMove} & $0.661$ & $0.265$& $2.371$\\
      & {PriSTI} & $0.401$ & \fst{$0.095$}& $1.488$\\
      & {DT + RP} & $1.024$ & $1.846$& $1.935$\\
      & {TW-DM} & \snd{$0.253$}& $0.139$& \fst{$0.834$}\\
      & {TW} & \fst{$0.018$}& \snd{$0.116$}& \snd{$1.084$}\\
    \hline
  \end{tabular}
  \end{adjustbox}
  \end{table}

  \noindent\textbf{Metrics.} We conduct experiments with three evaluation metrics to assess various aspects of seven different methods. They include Mean Squared Error (MSE) between the original trajectory and the recovered trajectory, which aims to measure how effectively models can recover broken trajectories. Normalized Dynamic Time Warping (NDTW) places emphasis on the shape of the trajectory rather than point-to-point comparison. Jensen–Shannon Divergence (JSD) is employed to compare the likelihood of two distributions, serving as a measure of the distribution of points between the recovered trajectories and the trajectories in the original dataset. It is important to note that lower values are preferable for all three metrics.

\subsection{Overall Performance}

We conduct comparative experiments among the proposed method and five baselines. 
\begin{itemize}
    \item \textbf{DeepMove} ~\cite{DeepMove} is an RNN-based method incorporating an attention mechanism, originally designed for trajectory forecasting. However, due to its reliance on sequential dependencies, it struggles to effectively capture complex relationships among points in trajectory recovery tasks, especially when the trajectories are sparse or irregular.
    \item \textbf{AttnMove} ~\cite{AttnMove} leverages both self-attention and cross-attention mechanisms to learn travel patterns from historical trajectories. It fuses intra- and inter-trajectory knowledge to enhance trajectory recovery, but may not be fully effective in scenarios where data is highly sparse or contains significant noise.
    \item \textbf{PriSTI} ~\cite{PriSTI} is a diffusion model developed for time-series imputation. We modified it to adapt to trajectory recovery tasks. While PriSTI occasionally achieves performance close to or exceeding TrajWeaver in certain metrics and datasets, it primarily serves as a strong baseline to validate our approach.
    \item \textbf{DT+RP} is a hybrid model that combines DiffTraj ~\cite{DiffTraj}, a UNet-based diffusion model for generating trajectories, with RePaint ~\cite{ImageInpainting}, a method designed for image inpainting. While DiffTraj generates trajectories without considering the given sparse trajectory, RePaint adapts diffusion-based generation techniques for recovery tasks. This combination aims to bridge the gap between generation and recovery but lacks the specific optimizations present in TrajWeaver.
    \item \textbf{TW} refers to TrajWeaver, the proposed method, which demonstrates superior performance across most metrics and datasets. This result highlights its robustness and effectiveness in recovering both pedestrian and vehicle trajectories in diverse scenarios.
    \item \textbf{TW-DM} is a variant of TrajWeaver that utilizes a standard diffusion model instead of the newly proposed \NewDM{}. While TW-DM achieves competitive results, its performance is slightly inferior to TW, underscoring the benefits of the state propagation mechanism introduced in \NewDM{}.
\end{itemize}

\subsection{Scalability Analysis}

  To evaluate the scalability of TrajWeaver across varying trajectory lengths and sparsity levels, we conducted experiments comparing it against strong baselines, including PriSTI and TrajWeaver without the state propagation diffusion model (SPDM). The experiments were performed using the Xi'an dataset, and the results are illustrated in Figure \ref{fig:Scalability}. In the three figures on the left, the sparsity level is fixed at 0.5, while the trajectory length is varied from 64 to 512. Conversely, in the three figures on the right, the trajectory length is held constant at 256, while the sparsity level is adjusted from 0.3 to 0.7. This setup allows us to comprehensively assess the model's performance under different conditions of data sparsity and sequence length.

  \begin{figure}[t]
    \centerline{\includegraphics[width=0.98\linewidth]{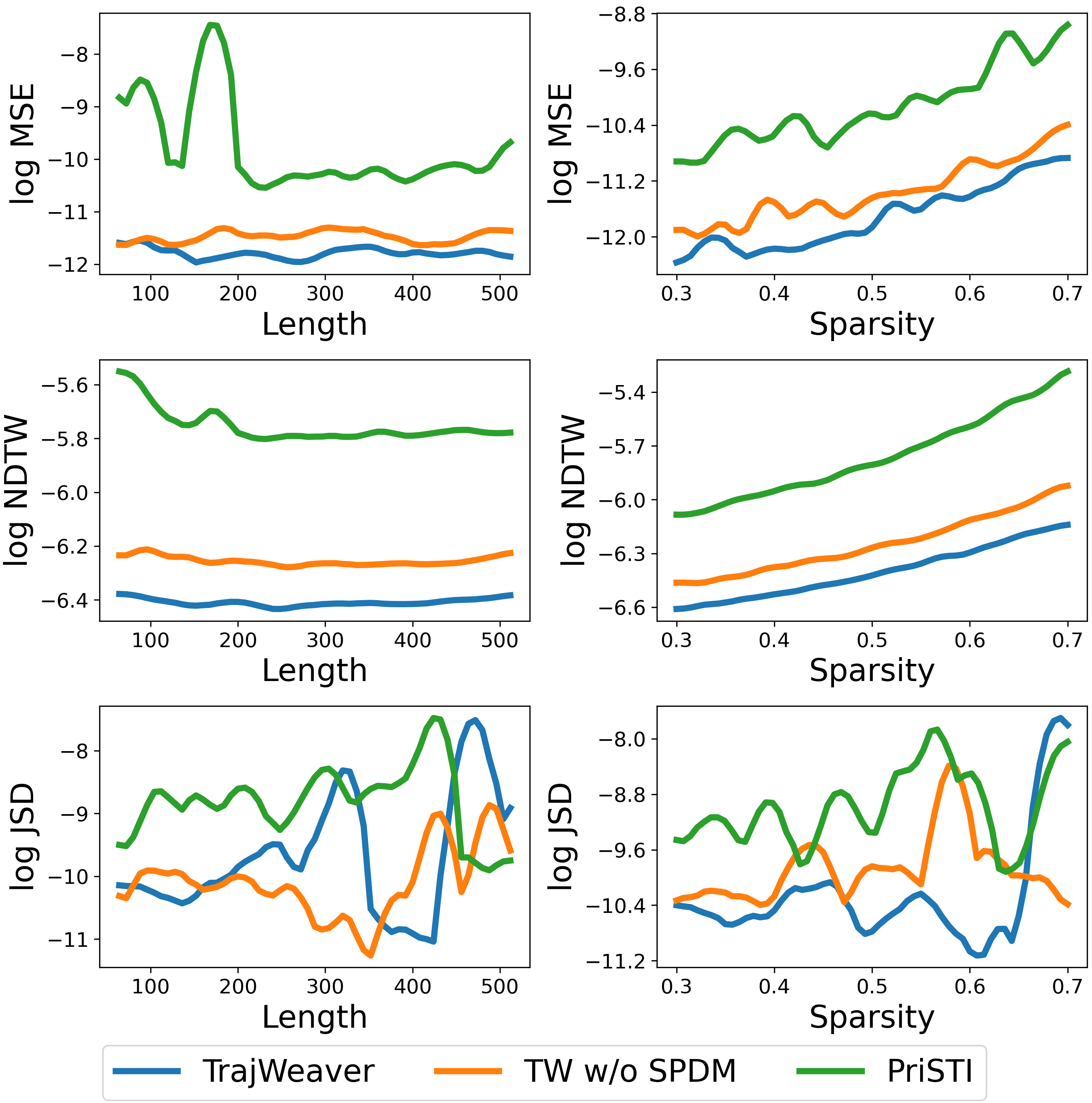}}
    \caption{The comparison among sparsity levels and trajectory lengths on Xi'an dataset.}
    \label{fig:Scalability}
  \end{figure}

\subsection{Efficiency Analysis}

  We conducted experiments to demonstrate that TrajWeaver can be both fast and accurate. DDIM is commonly used to accelerate diffusion models, although with some loss in recovery quality. Adding a state propagation pipeline into DDIM introduces 12\% more parameters to the model and obtains SP-DDIM. We compared the recovery time and quality of TrajWeaver with DDIM and SP-DDIM. The denoising time was recorded with a batch size of 100 on an NVIDIA GeForce RTX 3070 Ti Laptop GPU. As shown in Table \ref{tab:compare_time_quality}, the 21-step DDIM trades recovery quality for a 20.7 times speedup compared to the 500-step DDIM. In contrast, 21-step SP-DDIM achieves a 17.7 times speedup which is only 0.53 seconds slower than DDIM with the same number of steps, while still outperforming the 500-step DDIM in recovery quality by approximately 39.89\%.

\begin{table}[t]
    \caption{The comparison of recovery time and quality with and without state propagation pipeline.}
    \label{tab:compare_time_quality}
    \centering
    \begin{tabular}{|l||c|c|c|c|c|}
        \hline
         \TwoRow{Method} & \TwoRow{Steps} & Time & MSE & NDTW & JSD \\
         & & (s) & $\times 10^{-5}$ & $\times 10^{-3}$ & $\times 10^{-5}$ \\
         \hline
         \multirow{3}{*}{DDIM} & 500 & 65.47 & 1.78 & 2.01 & 3.94 \\
         & 51 & 7.31 & 1.79 & 2.06 & 12.58 \\
         & 21 & 3.16 & 2.01 & 2.31 & 28.71 \\
         \hline
         \multirow{3}{1cm}{\centering SP DDIM} & 500 & 80.20 & 0.86 & 1.75 & 3.69 \\
         & 51 & 8.58 & 0.94 & 1.75 & 6.78 \\
         & 21 & 3.69 & 1.07 & 1.92 & 4.05 \\
         \hline
    \end{tabular}
\end{table}

\begin{figure}[t]
\centering
\subfloat[State representations\label{fig:comp_state}]{\includegraphics[width=0.48\linewidth]{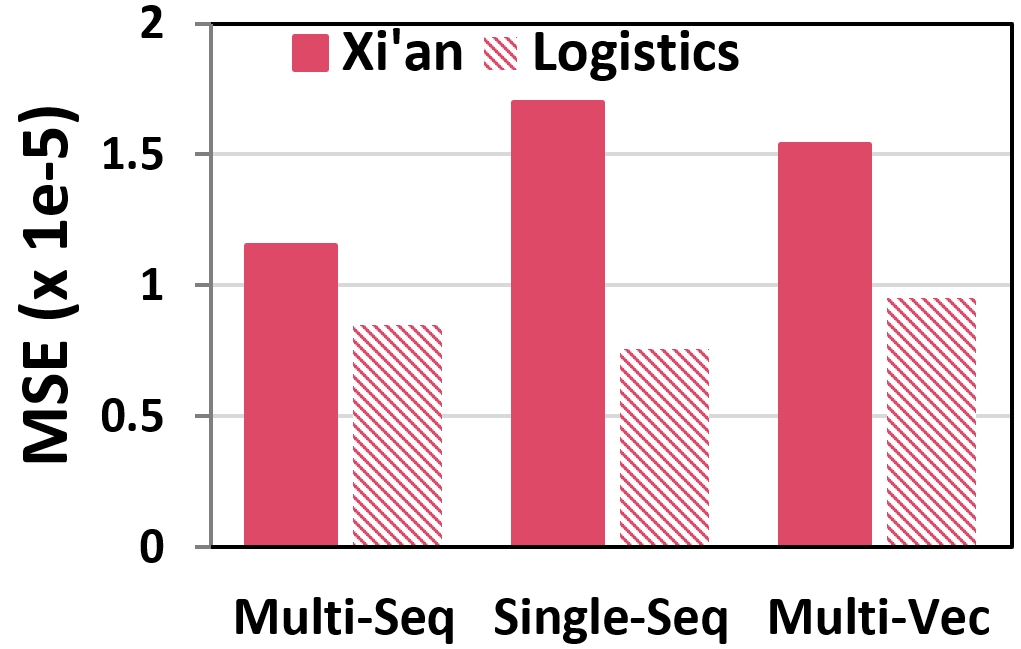}}
\subfloat[Fusion types\label{fig:comp_fusion}]{\includegraphics[width=0.48\linewidth]{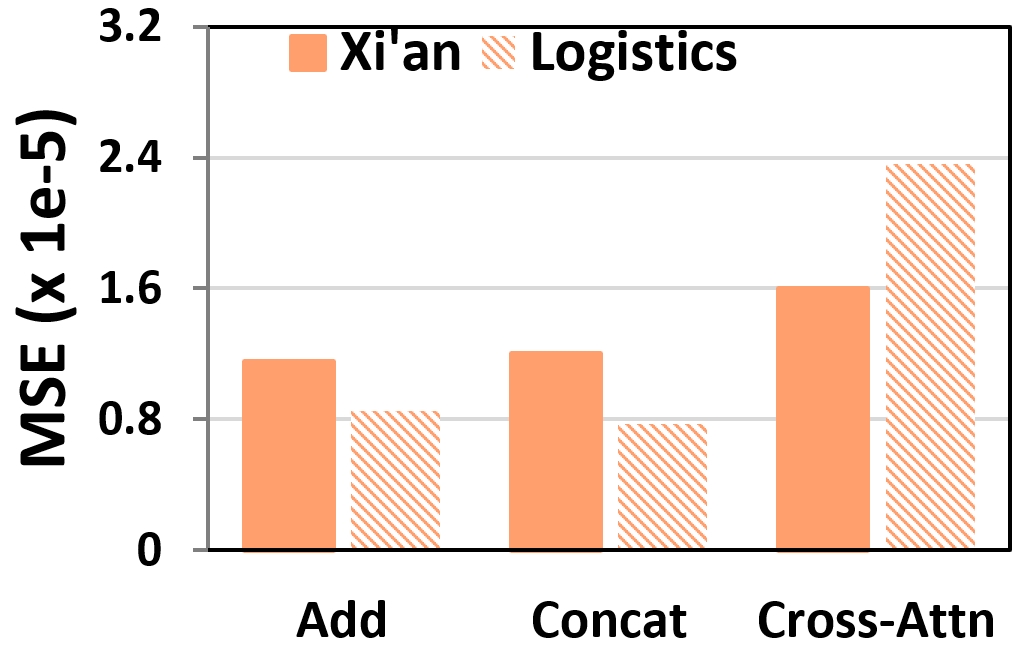}}\\

\subfloat[Steps in each iteration\label{fig:comp_step}]{\includegraphics[width=0.48\linewidth]{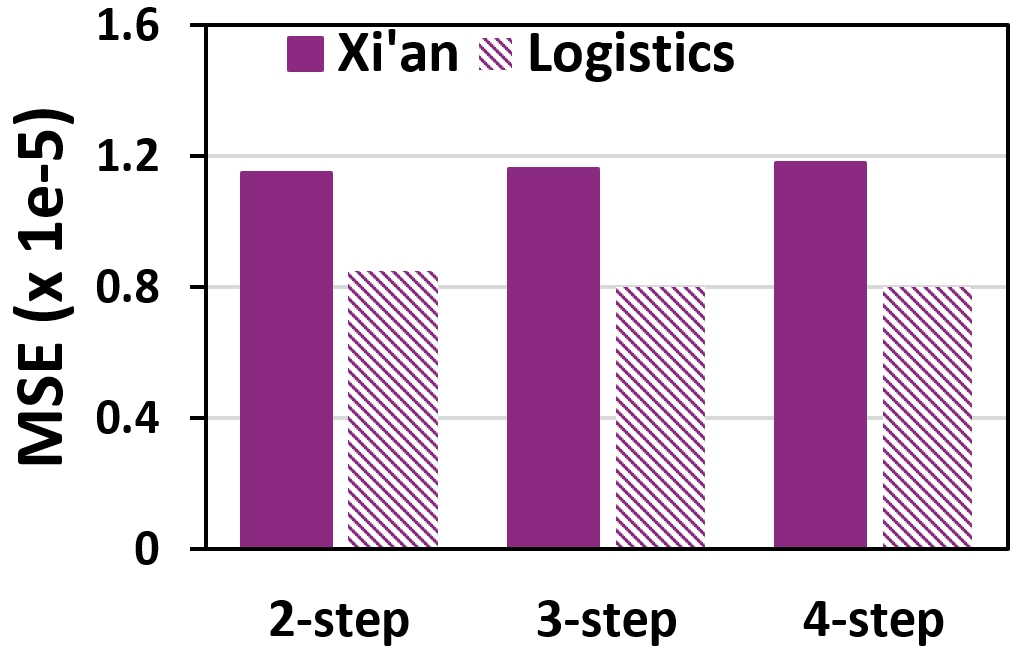}}
\subfloat[Batch managing\label{fig:comp_batching}]{\includegraphics[width=0.48\linewidth]{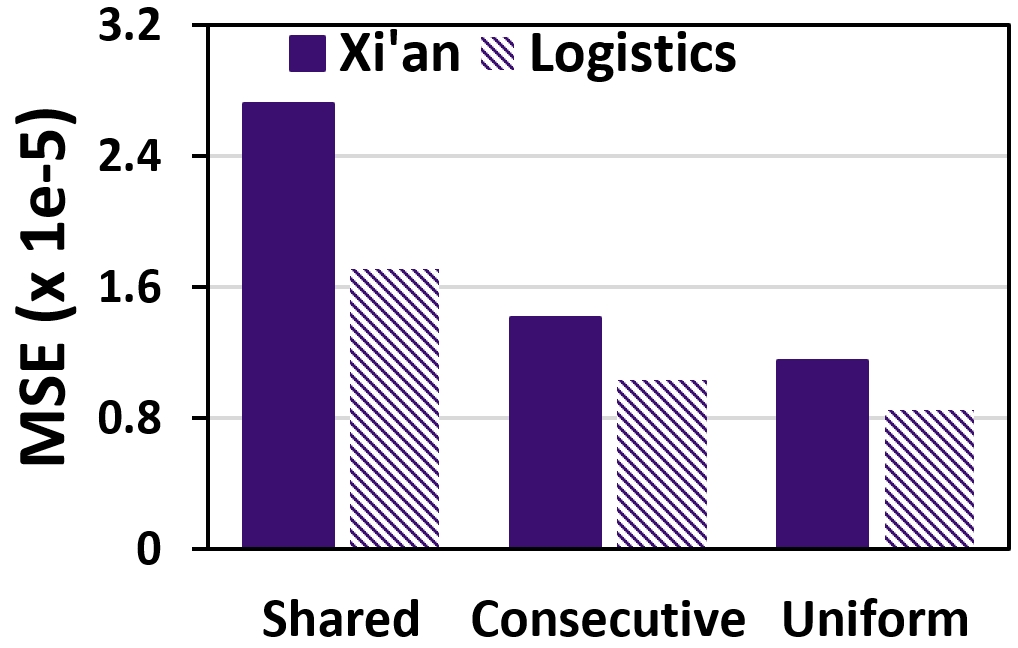}}

 \caption{The comparisons among: (a) Different state representations. (b) State fusion types. (c) Number of denoising steps trained in each training iteration. (d) Three batch managing algorithms introduced in Figure \ref{fig:batch_loading}.}
 \label{fig:comp_linkage_algo}
\end{figure}

  \subsection{Ablation Study}


  We analyze the effectiveness of different components in \NewDM{} structure and training algorithm using Xi'an and logistics dataset with trajectory length of 512 and sparsity of 0.5. The ablation studies show similar patterns for both dataset, indicating the model can generalize to both pedestrian and vehicle trajectories. Figure \ref{fig:comp_state} shows the outstanding performance of multi-scale sequential state design. Figure \ref{fig:comp_fusion} indicates that the state fusion can be either addition or concatenation, while cross attention is not very effective. We evaluated training 2, 3 and 4 adjacent steps in one iteration as shown in Figure \ref{fig:comp_step}. While training with more steps is not economic in practice as they cost too much time, computational power and memory. The result indicates that the number of denoising steps trained in a single training iteration has minimal effect on final performance. Finally, Figure \ref{fig:comp_batching} proves the robustness of the proposed batch managing algorithm.

  \subsection{Case Study}




  \begin{table}[t]
    \caption{The moving speed and distance estimations computed from trajectories after normalization.}
    \centering
    \begin{tabular}{|l||c|c|}
        \hline
        & Average Speed & Moving Distance \\
        & (m/s) & (km) \\
        \hline
        Ground Truth & 1 & 1\\
        \hline
        Sparse Traj & 0.6180 & 0.5788\\
        \hline
        Recovered Traj & 0.8252 & 0.8480 \\
        \hline
    \end{tabular}
    \label{tab:case_study_score}
    \end{table}

\begin{figure*}[th!]
  \centering
  \subfloat[Dense traj\label{fig:case_dense}]{\includegraphics[width=0.3\linewidth]{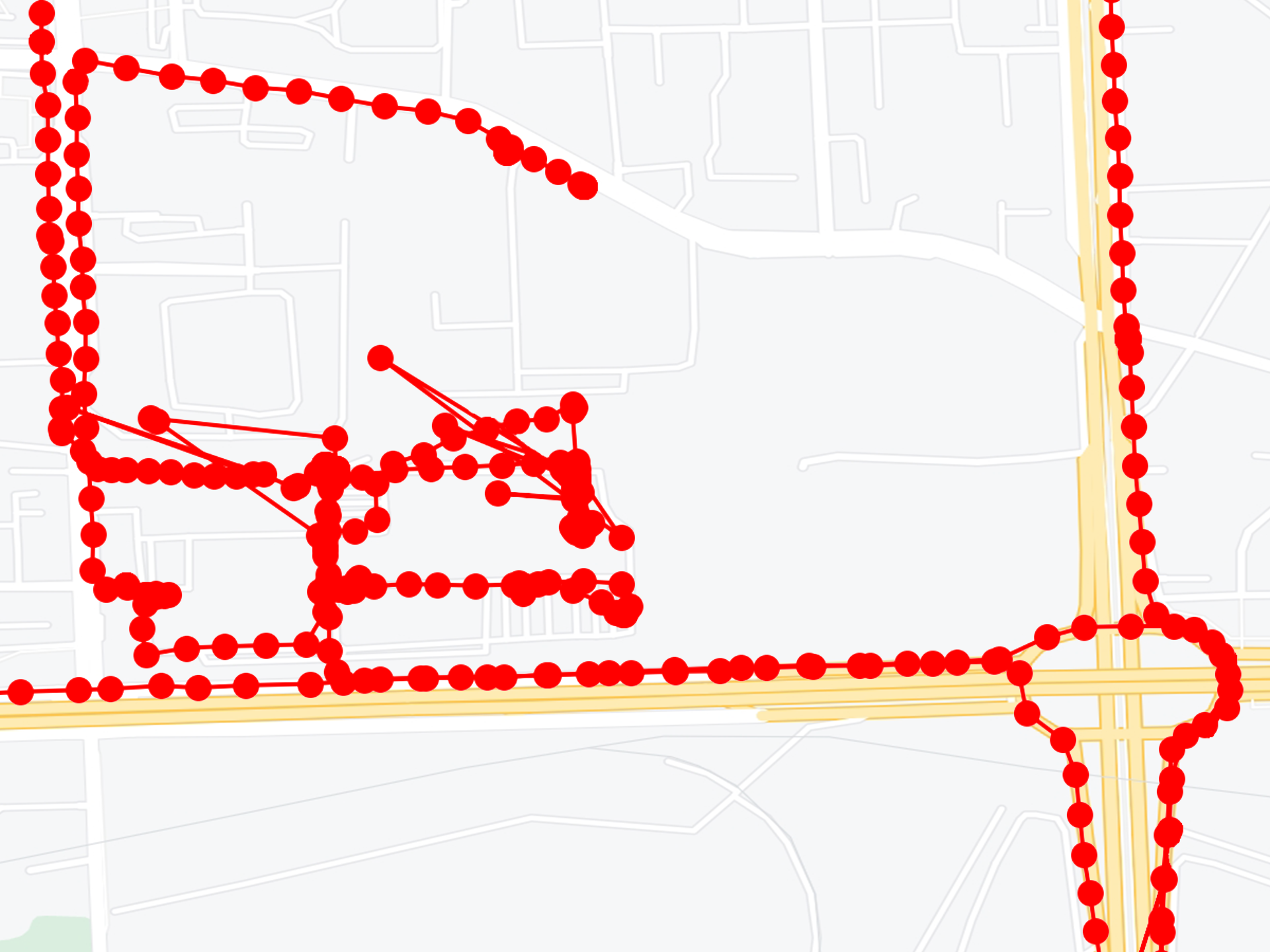}}
  \hfill
  \subfloat[Sparse traj\label{fig:case_sparse}]{\includegraphics[width=0.3\linewidth]{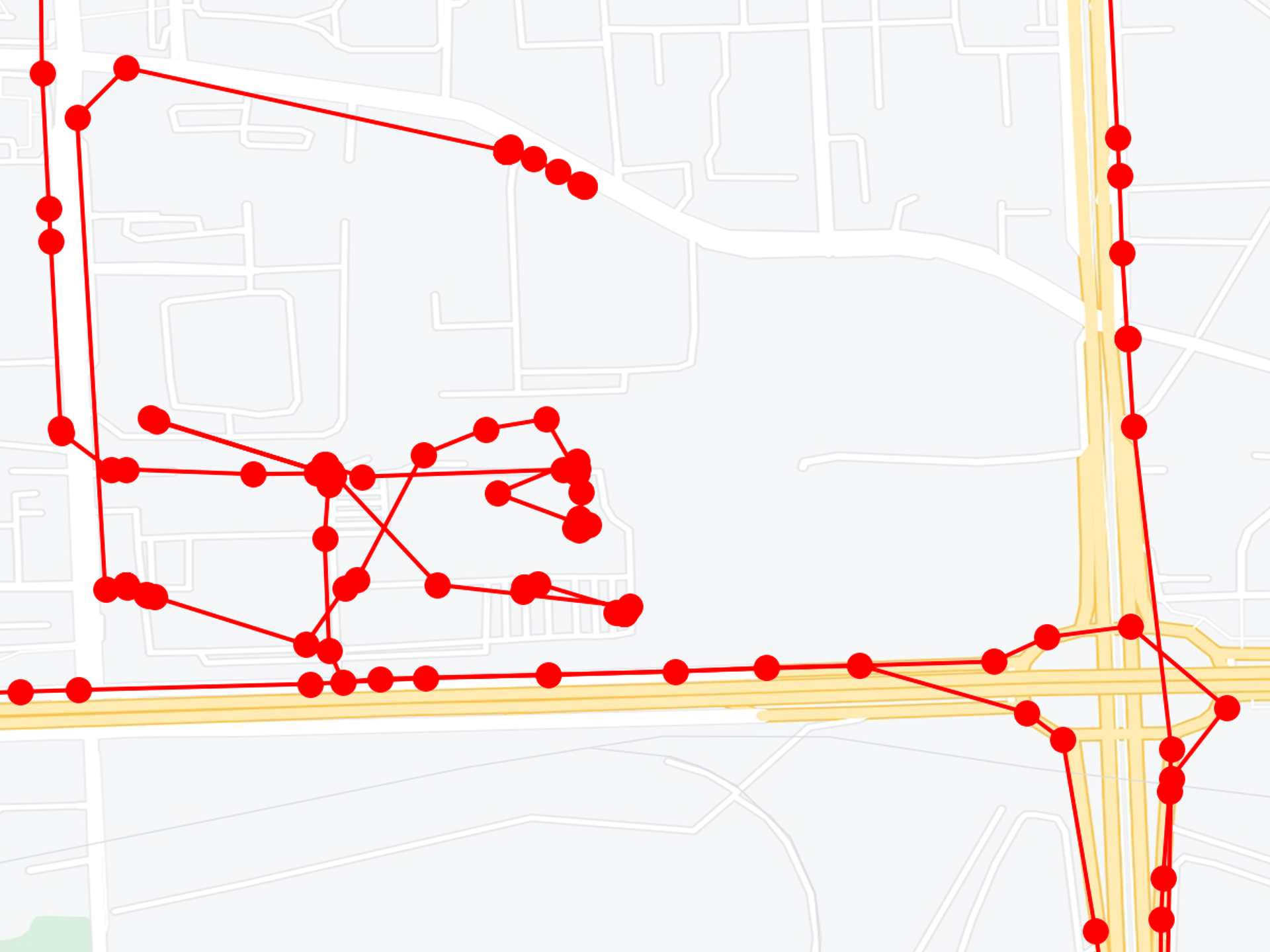}}
  \hfill
  \subfloat[Recovered traj\label{fig:case_recover_TW}]{\includegraphics[width=0.3\linewidth]{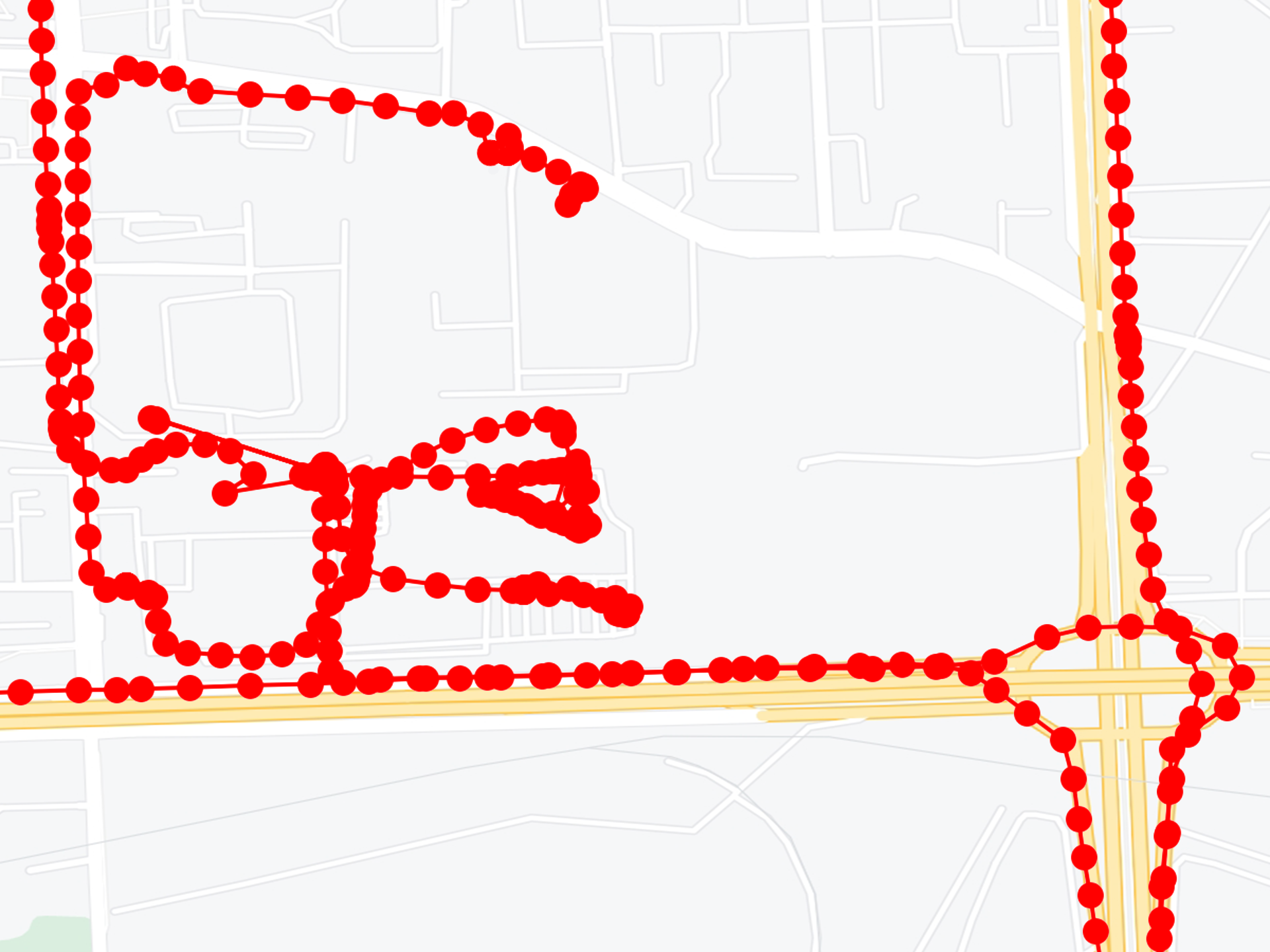}}
 \caption{The dataset of: (a) Dense trajectories as ground truth. (b) Trajectories with the same sparsity as usual logistics scenario. (c) Recovered trajectories using TrajWeaver.}
 \label{fig:case_study_visual}
\end{figure*}

A case study was conducted using trajectories of couriers in a logistics scenario. Monitoring the workload of couriers during last-mile delivery is necessary to ensure they are not overworked and are performing efficiently. Although workload can be measured from blood sugar levels and caloric expenditure, monitoring these indicators requires costly devices and poses privacy issues. Alternatively, accurate moving distance and speed can be used to estimate workload, which relies on dense trajectories. Since collecting and storing dense trajectories consume large amounts of power and storage, sparse trajectories are usually collected and later recovered to dense ones for analysis.

We collected very dense trajectories in a crowded apartment region. A subset was sampled from each dense trajectory according to a normal last-mile delivery sampling interval, resulting in only 30\% of the points remaining. TrajWeaver was trained to recover these sparse trajectories to dense ones. We used the moving distances and speed computed from the very dense trajectories as ground truth and then computed the estimation using the sparse and the recovered trajectories. For more direct comparison, we normalized the ground truth speed to 1 meter per second and the moving distance to 1 kilometer, and the results are shown in Table \ref{tab:case_study_score}. We observed that the estimations using sparse trajectories captured 60\% of the actual speed and distance, while the recovered trajectories increased this percentage to around 83\%. An illustration of the trajectory before and after recovery is shown in Figure \ref{fig:case_study_visual}.

\section{Conclusion}\label{sec:conclusion}
In conclusion, this paper presents TrajWeaver, a novel diffusion-based framework designed to address the challenges of trajectory recovery in urban environments. TrajWeaver leverages a new variant of the diffusion model, \NewDiffusionModel{} (\NewDM{}), which integrates multiple types of conditions to achieve efficient and high-quality recovery of sparse trajectory data. By introducing a state propagation mechanism, TrajWeaver enhances both the speed and accuracy of the recovery process, ensuring robust performance across different agent types, such as pedestrians and vehicles, even in crowded urban settings. The experimental results demonstrate TrajWeaver's scalability across varying levels of trajectory sparsity and length, and ablation studies confirm the effectiveness of its architectural design and specialized training algorithm. Moreover, the case study illustrates the practical applicability of the proposed method in real-world scenarios, showcasing its potential for deployment in diverse urban analytics tasks.

Looking ahead, there are several promising directions for future research. One potential avenue is to explore the generalization of the \NewDM{} structure beyond trajectory recovery to other domains, such as image or text generation, where efficient and high-quality recovery is also essential. Additionally, further improvements to TrajWeaver could be pursued by incorporating advancements from other diffusion techniques, such as latent diffusion models \cite{StableDiffusion}, to enhance performance or reduce computational costs. Future work might also investigate adaptive mechanisms for condition fusion that dynamically adjust to varying levels of data sparsity or explore the integration of real-time data streams to support dynamic, on-the-fly trajectory recovery in rapidly changing urban environments.


\bibliographystyle{IEEEtran}
\bibliography{Submission.bib}

\begin{thebibliography}{10}
\providecommand{\url}[1]{#1}
\csname url@samestyle\endcsname
\providecommand{\newblock}{\relax}
\providecommand{\bibinfo}[2]{#2}
\providecommand{\BIBentrySTDinterwordspacing}{\spaceskip=0pt\relax}
\providecommand{\BIBentryALTinterwordstretchfactor}{4}
\providecommand{\BIBentryALTinterwordspacing}{\spaceskip=\fontdimen2\font plus
\BIBentryALTinterwordstretchfactor\fontdimen3\font minus \fontdimen4\font\relax}
\providecommand{\BIBforeignlanguage}[2]{{%
\expandafter\ifx\csname l@#1\endcsname\relax
\typeout{** WARNING: IEEEtran.bst: No hyphenation pattern has been}%
\typeout{** loaded for the language `#1'. Using the pattern for}%
\typeout{** the default language instead.}%
\else
\language=\csname l@#1\endcsname
\fi
#2}}
\providecommand{\BIBdecl}{\relax}
\BIBdecl

\bibitem{Smallmap}
\BIBentryALTinterwordspacing
Z.~Hong, H.~Wang, Y.~Ding, G.~Wang, T.~He, and D.~Zhang, ``Smallmap: Low-cost community road map sensing with uncertain delivery behavior,'' \emph{Proc. ACM Interact. Mob. Wearable Ubiquitous Technol.}, vol.~8, no.~2, may 2024. [Online]. Available: \url{https://doi.org/10.1145/3659596}
\BIBentrySTDinterwordspacing

\bibitem{CNN_vehicle_cls}
\BIBentryALTinterwordspacing
S.~Dabiri, N.~Marković, K.~Heaslip, and C.~K. Reddy, ``A deep convolutional neural network based approach for vehicle classification using large-scale gps trajectory data,'' \emph{Transportation Research Part C: Emerging Technologies}, vol. 116, p. 102644, 2020. [Online]. Available: \url{https://www.sciencedirect.com/science/article/pii/S0968090X20305593}
\BIBentrySTDinterwordspacing

\bibitem{trans_mode_identify}
J.~Li, X.~Pei, X.~Wang, D.~Yao, Y.~Zhang, and Y.~Yue, ``Transportation mode identification with gps trajectory data and gis information,'' \emph{Tsinghua Science and Technology}, vol.~26, no.~4, pp. 403--416, 2021.

\bibitem{ImageInpainting}
\BIBentryALTinterwordspacing
A.~Lugmayr, M.~Danelljan, A.~Romero, F.~Yu, R.~Timofte, and L.~V. Gool, ``Repaint: Inpainting using denoising diffusion probabilistic models,'' \emph{CoRR}, vol. abs/2201.09865, 2022. [Online]. Available: \url{https://arxiv.org/abs/2201.09865}
\BIBentrySTDinterwordspacing

\bibitem{PriSTI}
\BIBentryALTinterwordspacing
M.~Liu, H.~Huang, H.~Feng, L.~Sun, B.~Du, and Y.~Fu, ``Pristi: A conditional diffusion framework for spatiotemporal imputation,'' in \emph{2023 IEEE 39th International Conference on Data Engineering (ICDE)}.\hskip 1em plus 0.5em minus 0.4em\relax IEEE, Apr. 2023. [Online]. Available: \url{http://dx.doi.org/10.1109/ICDE55515.2023.00150}
\BIBentrySTDinterwordspacing

\bibitem{VAE}
D.~P. Kingma and M.~Welling, ``Auto-encoding variational bayes,'' 2022.

\bibitem{TrajVAE}
\BIBentryALTinterwordspacing
X.~Chen, J.~Xu, R.~Zhou, W.~Chen, J.~Fang, and C.~Liu, ``Trajvae: A variational autoencoder model for trajectory generation,'' \emph{Neurocomputing}, vol. 428, pp. 332--339, 2021. [Online]. Available: \url{https://www.sciencedirect.com/science/article/pii/S0925231220312017}
\BIBentrySTDinterwordspacing

\bibitem{GAN}
I.~J. Goodfellow, J.~Pouget-Abadie, M.~Mirza, B.~Xu, D.~Warde-Farley, S.~Ozair, A.~Courville, and Y.~Bengio, ``Generative adversarial networks,'' 2014.

\bibitem{TrajGANContinuous}
\BIBentryALTinterwordspacing
W.~Jiang, W.~X. Zhao, J.~Wang, and J.~Jiang, ``Continuous trajectory generation based on two-stage gan,'' \emph{Proceedings of the AAAI Conference on Artificial Intelligence}, vol.~37, no.~4, pp. 4374--4382, Jun. 2023. [Online]. Available: \url{https://ojs.aaai.org/index.php/AAAI/article/view/25557}
\BIBentrySTDinterwordspacing

\bibitem{ddpm_beat_gan_topology}
F.~Maz{\'e} and F.~Ahmed, ``Diffusion models beat gans on topology optimization,'' in \emph{Proceedings of the AAAI conference on artificial intelligence}, vol.~37, 2023, pp. 9108--9116.

\bibitem{text-to-image-vqvae}
S.~Gu, D.~Chen, J.~Bao, F.~Wen, B.~Zhang, D.~Chen, L.~Yuan, and B.~Guo, ``Vector quantized diffusion model for text-to-image synthesis,'' in \emph{Proceedings of the IEEE/CVF conference on computer vision and pattern recognition}, 2022, pp. 10\,696--10\,706.

\bibitem{Sora}
Y.~Liu, K.~Zhang, Y.~Li, Z.~Yan, C.~Gao, R.~Chen, Z.~Yuan, Y.~Huang, H.~Sun, J.~Gao, L.~He, and L.~Sun, ``Sora: A review on background, technology, limitations, and opportunities of large vision models,'' 2024.

\bibitem{StableDiffusion}
\BIBentryALTinterwordspacing
R.~Rombach, A.~Blattmann, D.~Lorenz, P.~Esser, and B.~Ommer, ``High-resolution image synthesis with latent diffusion models,'' \emph{CoRR}, vol. abs/2112.10752, 2021. [Online]. Available: \url{https://arxiv.org/abs/2112.10752}
\BIBentrySTDinterwordspacing

\bibitem{DDPM}
\BIBentryALTinterwordspacing
J.~Ho, A.~Jain, and P.~Abbeel, ``Denoising diffusion probabilistic models,'' \emph{CoRR}, vol. abs/2006.11239, 2020. [Online]. Available: \url{https://arxiv.org/abs/2006.11239}
\BIBentrySTDinterwordspacing

\bibitem{DDIM}
\BIBentryALTinterwordspacing
J.~Song, C.~Meng, and S.~Ermon, ``Denoising diffusion implicit models,'' \emph{CoRR}, vol. abs/2010.02502, 2020. [Online]. Available: \url{https://arxiv.org/abs/2010.02502}
\BIBentrySTDinterwordspacing

\bibitem{DPMSolver}
C.~Lu, Y.~Zhou, F.~Bao, J.~Chen, C.~Li, and J.~Zhu, ``Dpm-solver: A fast ode solver for diffusion probabilistic model sampling in around 10 steps,'' 2022.

\bibitem{PNDM}
\BIBentryALTinterwordspacing
L.~Liu, Y.~Ren, Z.~Lin, and Z.~Zhao, ``Pseudo numerical methods for diffusion models on manifolds,'' 2022. [Online]. Available: \url{https://arxiv.org/abs/2202.09778}
\BIBentrySTDinterwordspacing

\bibitem{DiffPruning}
\BIBentryALTinterwordspacing
G.~Fang, X.~Ma, and X.~Wang, ``Structural pruning for diffusion models,'' 2023. [Online]. Available: \url{https://arxiv.org/abs/2305.10924}
\BIBentrySTDinterwordspacing

\bibitem{one_step_diffusion}
\BIBentryALTinterwordspacing
T.~Yin, M.~Gharbi, R.~Zhang, E.~Shechtman, F.~Durand, W.~T. Freeman, and T.~Park, ``One-step diffusion with distribution matching distillation,'' 2023. [Online]. Available: \url{https://arxiv.org/abs/2311.18828}
\BIBentrySTDinterwordspacing

\bibitem{DeepCache}
X.~Ma, G.~Fang, and X.~Wang, ``Deepcache: Accelerating diffusion models for free,'' 2023.

\bibitem{HumanTrajCompletionTrans}
J.~Ma, C.~Yang, S.~Mao, J.~Zhang, S.~C. Periaswamy, and J.~Patton, ``Human trajectory completion with transformers,'' in \emph{ICC 2022 - IEEE International Conference on Communications}, 2022, pp. 3346--3351.

\bibitem{TaxiTrajRec}
F.~Hou, X.~Lan, J.~Chen, Y.~Dong, X.~Zhuang, and J.~Wu, ``A sparse taxi trajectory data recovery and calibrate algorithm,'' in \emph{2021 China Automation Congress (CAC)}, 2021, pp. 5414--5419.

\bibitem{MTrajRec}
\BIBentryALTinterwordspacing
H.~Ren, S.~Ruan, Y.~Li, J.~Bao, C.~Meng, R.~Li, and Y.~Zheng, ``Mtrajrec: Map-constrained trajectory recovery via seq2seq multi-task learning,'' in \emph{Proceedings of the 27th ACM SIGKDD Conference on Knowledge Discovery \& Data Mining}, ser. KDD '21.\hskip 1em plus 0.5em minus 0.4em\relax New York, NY, USA: Association for Computing Machinery, 2021, p. 1410–1419. [Online]. Available: \url{https://doi.org/10.1145/3447548.3467238}
\BIBentrySTDinterwordspacing

\bibitem{RNTrajRec}
Y.~Chen, H.~Zhang, W.~Sun, and B.~Zheng, ``Rntrajrec: Road network enhanced trajectory recovery with spatial-temporal transformer,'' 2022.

\bibitem{TrajRecCalibKF}
J.~Wang, N.~Wu, X.~Lu, W.~X. Zhao, and K.~Feng, ``Deep trajectory recovery with fine-grained calibration using kalman filter,'' \emph{IEEE Transactions on Knowledge and Data Engineering}, vol.~33, no.~3, pp. 921--934, 2021.

\bibitem{AttnMove}
\BIBentryALTinterwordspacing
T.~Xia, Y.~Li, Y.~Qi, J.~Feng, F.~Xu, F.~Sun, D.~Guo, and D.~Jin, ``History-enhanced and uncertainty-aware trajectory recovery via attentive neural network,'' \emph{ACM Trans. Knowl. Discov. Data}, vol.~18, no.~3, dec 2023. [Online]. Available: \url{https://doi.org/10.1145/3615660}
\BIBentrySTDinterwordspacing

\bibitem{Transformer}
\BIBentryALTinterwordspacing
A.~Vaswani, N.~Shazeer, N.~Parmar, J.~Uszkoreit, L.~Jones, A.~N. Gomez, L.~Kaiser, and I.~Polosukhin, ``Attention is all you need,'' \emph{CoRR}, vol. abs/1706.03762, 2017. [Online]. Available: \url{http://arxiv.org/abs/1706.03762}
\BIBentrySTDinterwordspacing

\bibitem{TrajBert}
J.~Si, J.~Yang, Y.~Xiang, H.~Wang, L.~Li, R.~Zhang, B.~Tu, and X.~Chen, ``Trajbert: Bert-based trajectory recovery with spatial-temporal refinement for implicit sparse trajectories,'' \emph{IEEE Transactions on Mobile Computing}, vol.~23, no.~5, pp. 4849--4860, 2024.

\bibitem{BERT}
\BIBentryALTinterwordspacing
J.~Devlin, M.~Chang, K.~Lee, and K.~Toutanova, ``{BERT:} pre-training of deep bidirectional transformers for language understanding,'' \emph{CoRR}, vol. abs/1810.04805, 2018. [Online]. Available: \url{http://arxiv.org/abs/1810.04805}
\BIBentrySTDinterwordspacing

\bibitem{CSDI}
Y.~Tashiro, J.~Song, Y.~Song, and S.~Ermon, ``Csdi: Conditional score-based diffusion models for probabilistic time series imputation,'' 2021.

\bibitem{SSSD}
J.~M.~L. Alcaraz and N.~Strodthoff, ``Diffusion-based time series imputation and forecasting with structured state space models,'' 2023.

\bibitem{DiffTraj}
Y.~Zhu, Y.~Ye, S.~Zhang, X.~Zhao, and J.~J.~Q. Yu, ``Difftraj: Generating gps trajectory with diffusion probabilistic model,'' 2023.

\bibitem{UNet}
\BIBentryALTinterwordspacing
O.~Ronneberger, P.~Fischer, and T.~Brox, ``U-net: Convolutional networks for biomedical image segmentation,'' \emph{CoRR}, vol. abs/1505.04597, 2015. [Online]. Available: \url{http://arxiv.org/abs/1505.04597}
\BIBentrySTDinterwordspacing

\bibitem{diff_rn_traj}
\BIBentryALTinterwordspacing
T.~Wei, Y.~Lin, S.~Guo, Y.~Lin, Y.~Huang, C.~Xiang, Y.~Bai, M.~Ya, and H.~Wan, ``Diff-rntraj: A structure-aware diffusion model for road network-constrained trajectory generation,'' 2024. [Online]. Available: \url{https://arxiv.org/abs/2402.07369}
\BIBentrySTDinterwordspacing

\bibitem{PTQD}
\BIBentryALTinterwordspacing
Y.~He, L.~Liu, J.~Liu, W.~Wu, H.~Zhou, and B.~Zhuang, ``Ptqd: Accurate post-training quantization for diffusion models,'' 2023. [Online]. Available: \url{https://arxiv.org/abs/2305.10657}
\BIBentrySTDinterwordspacing

\bibitem{DeepMove}
\BIBentryALTinterwordspacing
J.~Feng, Y.~Li, C.~Zhang, F.~Sun, F.~Meng, A.~Guo, and D.~Jin, ``Deepmove: Predicting human mobility with attentional recurrent networks,'' in \emph{Proceedings of the 2018 World Wide Web Conference}, ser. WWW '18.\hskip 1em plus 0.5em minus 0.4em\relax Republic and Canton of Geneva, CHE: International World Wide Web Conferences Steering Committee, 2018, p. 1459–1468. [Online]. Available: \url{https://doi.org/10.1145/3178876.3186058}
\BIBentrySTDinterwordspacing

\end{thebibliography}




\end{document}